\documentclass[letterpaper]{article} 
\usepackage{aaai2027}  
\usepackage[hyphens]{url}  
\usepackage{graphicx} 
\usepackage{natbib}  
\usepackage{caption} 
\usepackage{algorithm}
\usepackage{algorithmic}

\usepackage{newfloat}
\usepackage{listings}
\DeclareCaptionStyle{ruled}{labelfont=normalfont,labelsep=colon,strut=off} 
\floatstyle{ruled}
\newfloat{listing}{tb}{lst}{}
\floatname{listing}{Listing}

\usepackage{booktabs}

\usepackage{times}  
\usepackage{helvet}  
\usepackage{courier}  
\usepackage[hyphens]{url}  
\usepackage{graphicx} 
\usepackage{natbib}  
\usepackage{caption} 
\usepackage{amssymb}
\usepackage{subcaption}
\newcommand{\shortname}{\textit{LeanMem}}
\usepackage{paralist}
\usepackage{graphicx}
\usepackage{enumitem}
\usepackage{multirow}
\usepackage[table]{xcolor}
\definecolor{rowbase}{RGB}{242,242,242} 
\definecolor{rowrag}{RGB}{224,238,250}  
\definecolor{rowft}{RGB}{228,244,228}   
\definecolor{rowmem}{RGB}{241,231,247}  
\definecolor{rowours}{RGB}{255,246,212} 

\usepackage{algorithm}
\usepackage{algorithmic}
\usepackage{adjustbox}
\usepackage{booktabs}
\usepackage[table]{xcolor}
\usepackage{amsmath}
\usepackage{newfloat}
\usepackage{listings}
\nocopyright 

\title{LeanMem: Simple and Efficient Long-Term Memory for LLM Agents}
\author{
    Yuxin Liao,
    Le Wu\thanks{Corresponding author.},
    Min Hou,
    Hao Liu,
    Han Wu,
    Zishu Wang
}
\affiliations{

    Hefei University of Technology\\
    yuxinliao314@gmail.com, lewu.ustc@gmail.com, hmhoumin@gmail.com, hliu7879@163.com, ustcwuhan@gmail.com, zishuwang@mail.hfut.edu.cn
}

\begin{document}

\maketitle

\begin{abstract}

Long-term memory is essential for LLM-based agents to sustain interactions and reliably leverage distant history. However, existing memory systems typically process heterogeneous dialogue content through a uniform summarization and retrieval pipeline, leading to either excessive token consumption or irreversible loss of fine-grained evidence. We argue that historical dialogue content should be handled differently according to its compressibility, temporal dynamics, and fidelity requirements.
Based on this insight, we propose LeanMem, a lightweight long-term memory framework. LeanMem first filters out low-value content, then stores informative segments as compact profile memory, temporally structured event memory, or source-grounded record memory, depending on the nature of the information. During maintenance, only dynamically evolving event memories are selectively updated, avoiding redundant consolidation of stable profiles and immutable records. During inference, \shortname~dynamically selects memory types and allocates retrieval budgets according to query-specific evidence demands, assembling relevant evidence on demand. On LoCoMo and LongMemEval-S with GPT-4.1-mini and Qwen3-8B, LeanMem improves accuracy over the strongest memory-based baseline in every setting, by up to 15.1 points, at the lowest or near-lowest construction cost, inference tokens, and latency. The code and datasets are included in the supplementary materials.
\end{abstract}

\section{Introduction}
Large language model (LLM)-based agents have shown strong capabilities across diverse tasks~\cite{li2024survey,ferrag2026llm}. However, their ability to support long-running interactions is limited by fixed context windows and unreliable access to distant history. To address this, memory~\cite{memorysurvey1,memorysurvey2,memorysurvey3} has become a pivot mechanism for LLM-based agents, allowing them to maintain a persistent state over historical interactions. It enables agents to retain user-specific knowledge, track evolving information, and retrieve relevant evidence for future decision-making~\cite{liu2026exploratory,steam, banerjee2026apex}.

\begin{figure}[t]
    \centering
    \includegraphics[
        width=\linewidth,
        height=0.28\textheight,
        keepaspectratio
    ]{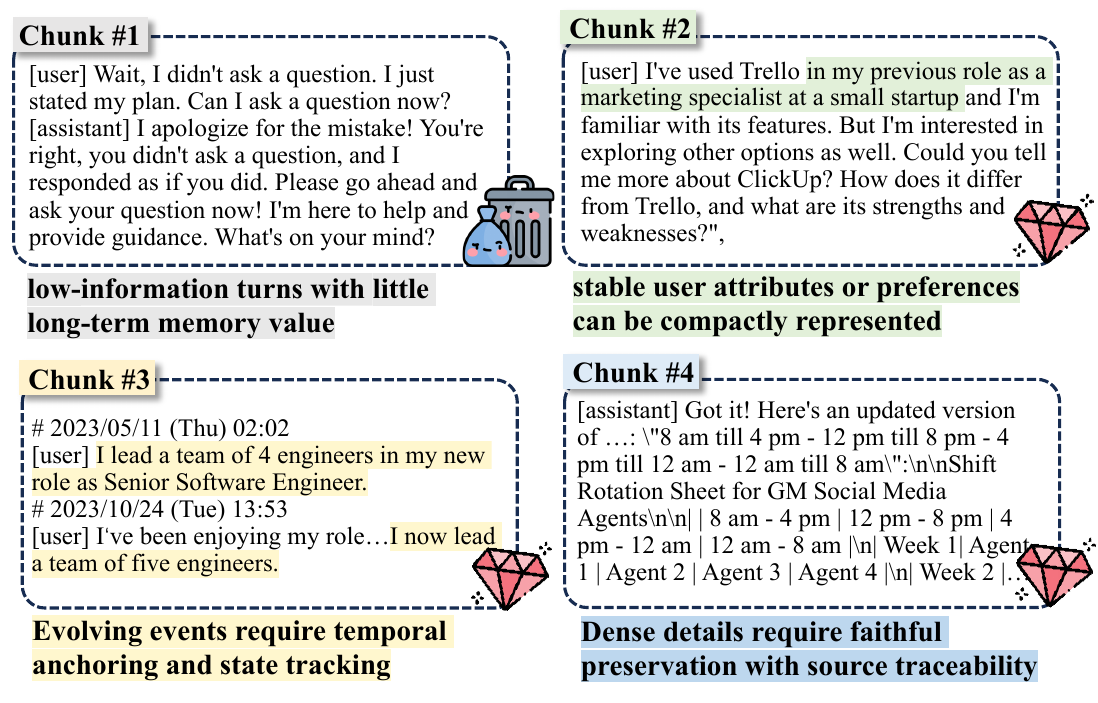}
    \caption{
Dialogue information differs in compressibility and storage requirements.
}
    \label{fig:intro}
\end{figure}

Recent studies~\cite{A-mem,cam,zhong2026hingemem} have extensively explored the design of agent memory, covering the full pipeline from extracting information out of raw interactions, to constructing and evolving memory units, and ultimately retrieving relevant evidence for downstream reasoning and response generation. 
Despite these advances, \textbf{memory systems still suffer from significant token consumption and information loss}~\cite{simplemem,lightmem}.
In this pipeline, memory construction is usually the first major source of cost~\cite{huang2026rethinking}. For each incoming dialogue chunk, existing systems often process full user-assistant dialogue chunks to an LLM to identify useful content, summarize it into memory entries, and update the existing memory bank. This eager construction strategy keeps memories immediately available, but it repeatedly invokes LLMs over long interaction histories and therefore consumes substantial construction tokens. At the same time, reducing raw interactions into compact memories can discard fine-grained details that are difficult to recover later. \textbf{This creates a tension between efficiency and fidelity}: processing more context improves the chance of preserving useful evidence, but it increases token cost; compressing more aggressively reduces cost, but it may lose information needed by future queries.

Recent studies have tried to address this tension from different sides. (1) Lightweight memory construction methods reduce token consumption by filtering redundant dialogue, compressing input before memory writing, or delaying consolidation to an offline stage~\cite{lightmem,simplemem}. These methods improve efficiency, but they mainly reduce how much information enters the memory pipeline. Once useful information is compressed or filtered out, fine-grained evidence may still be difficult to recover for future queries. (2) In contrast, retrieval-side methods~\cite{zhang2026memsearch,kim2025pre} improve fidelity by using iterative retrieval, or multi-hop reasoning to collect more evidence at inference time. These methods can improve answer accuracy, but they introduce additional token consumption and latency during retrieval and response generation. Therefore, existing approaches often improve one side of the efficiency-fidelity trade-off at the cost of the other.

We address the efficiency and fidelity issues from a different perspective. \textbf{Our key observation is that dialogue information differs in compressibility and storage requirements.} Not every dialogue segment should be recorded indiscriminately. Some segments contain little useful information and can be safely ignored. Some contain stable user attributes or preferences, which can be stored in a compact structured form. Some describe events that change over time, which require temporal anchors and state tracking. Others contain dense details, such as lists, plans, or step-by-step discussions, where aggressive summarization may lose important evidence. Therefore, an efficient memory system should not apply the same construction strategy to all dialogue content. It should first decide whether a segment should be stored, and then choose a storage form that matches its information type and fidelity requirement. As shown in Fig.~\ref{fig:intro}, stable profiles, evolving events, and detail-intensive records require different degrees of compression and preservation.

Based on this idea, we propose \shortname, an extremely lightweight long-term memory framework for LLM-based agents. \shortname~treats memory construction as a controlled writing process. We develop a three-stage pipeline: (1) \textbf{Controlled Memory Writing}: Given a dialogue segment, \shortname~first decides whether the segment should enter memory. Segments with low information value are ignored to avoid unnecessary construction cost. For useful segments, \shortname~uses a lightweight scheduler to classify the information type and assign a suitable storage strategy. Stable user attributes and preferences are stored as compact profile memory. State-changing information is stored as event memory with temporal anchors and state descriptions. Detail-intensive content is stored as record memory with lightweight indices and pointers to the original dialogue, so that fine-grained evidence can be traced back when needed. (2) \textbf{Selective Memory Evolution}: we further perform memory evolution only for event memory. It groups related events across sessions and merges them into temporal state representations. This moves part of the evidence integration required by multi-hop temporal queries from query time to the memory construction stage.(3) \textbf{Adaptive Evidence Composition}: At inference time, we perform adaptive evidence composition by selecting memory types and retrieval budgets according to the query. In this way, \shortname~reduces redundant memory writing, preserves important details, and avoids expensive multi-hop retrieval for every query.

We evaluate \shortname~on two long-term conversational question answering benchmarks, LongMemEval-S and LoCoMo. \shortname~consistently improves answer accuracy while substantially reducing token consumption. Compared with A-Mem, it improves Accuracy by up to 20.40 percentage points, reduces memory construction tokens by up to 17.24$\times$, and reduces inference tokens by up to 8.41$\times$. These results demonstrate that \shortname~achieves a stronger effectiveness--efficiency balance by representing conversational information according to its compressibility, temporal dynamics, and fidelity requirements, and retrieving evidence according to query-specific demands.


\begin{figure*}[t]
\setlength{\abovecaptionskip}{0.1cm}
  \centering
  \includegraphics[width=\textwidth]{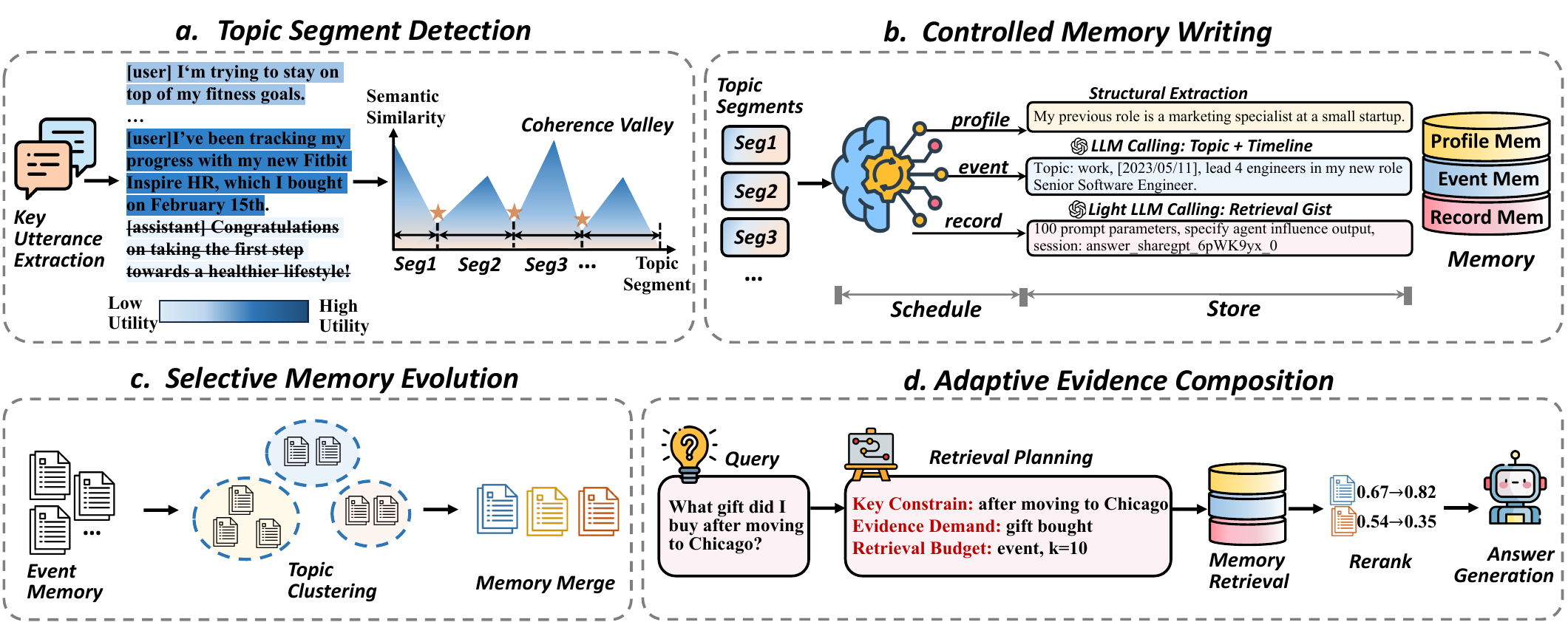}
    \caption{
    The architecture of \shortname.
    (a) Topic Segmentation filters key utterances and groups them into topically coherent segments.
    (b) Controlled Memory Writing routes each segment to profile, event, or record memory according to its compression, temporal, and fidelity requirements.
    (c) Selective Memory Evolution buffers and incrementally updates event memory through localized offline consolidation.
    (d) Adaptive Evidence Composition plans retrieval and assembles sufficient evidence under a query-conditioned budget.
    }
  \label{fig:framework}
\vspace{-10pt}

\end{figure*}

\section{Preliminaries}

\paragraph{Long-Term Memory Pipeline.}

A conversational agent interacts with a user across multiple sessions, producing a dialogue history $\mathcal{H}={c_i}*{i=1}^{C}$, where each session $c_i={x*{i,j}}_{j=1}^{T_i}$ contains $T_i$ user--assistant turns. An explicit memory system extracts valuable information from $\mathcal{H}$ into an external memory bank $\mathcal{M}$ to support future queries.

Most agent memory systems involve two stages: memory construction and memory usage. During construction, raw dialogues are transformed into retrievable entries and evolved as interactions accumulate. During usage, a query $q$ retrieves supporting evidence from $\mathcal{M}$ for answer generation. Some systems further iterate retrieval and reflection until sufficient evidence is obtained. Representative LLM agent memory systems are reviewed in Appendix~A.

\section{Method}

\subsection{Overview}
The central proposal of \shortname~is that conversational information differs in compressibility, temporal dynamics, and fidelity requirements. However, existing memory systems typically process dialogue history using a uniform compression strategy, resulting in either insufficient compression or irreversible information loss.

As illustrated in Fig.~\ref{fig:framework}, \shortname~addresses this mismatch through three coordinated stages spanning memory construction, maintenance, and retrieval. First, \textbf{Controlled Memory Writing} partitions dialogue history into coherent segments and stores them as compact profile memory, evolving event memory, or source-grounded record memory according to their compressibility, temporal dynamics, and fidelity requirements. Second, \textbf{Selective Memory Evolution} applies deferred and localized updates only to event memory, avoiding unnecessary consolidation of stable profiles and immutable records. Third, \textbf{Adaptive Evidence Composition} identifies query-specific evidence demands, selects relevant memory types, and allocates retrieval budgets to assemble sufficient evidence. Through these stages, \shortname~reduces long-term memory cost while preserving the evidence required for reliable downstream reasoning.

\subsection{Controlled Memory Writing}
Given a dialogue history $\mathcal{H}$, Controlled Memory Writing first filters it into key utterances $\mathcal{U}$, which are further grouped into semantically coherent topic segments $\mathcal{S}=\{S_i\}_{i=1}^{N}$, where $N$ is the number of detected segments. Each segment $S_i$ serves as the basic decision unit for write scheduling, producing a routing result $r_i$ that guides its subsequent memory materialization.

\noindent\textbf{$\bullet$ Key Utterance Filtering. }
\shortname~first applies a rule-based filter to remove structurally low-information turns, including greetings, acknowledgements, and restatements. In asymmetric user--assistant interactions, user utterances are treated as the primary routing anchors because they typically introduce user-specific requirements, preferences, and state changes, whereas assistant responses mainly elaborate on the current request. Assistant utterances are therefore not independently retained as routing anchors in $\mathcal{U}$, while their contextual information remains accessible from the original dialogue. This preprocessing produces a cleaner utterance sequence and reduces subsequent LLM input.

\noindent\textbf{$\bullet$ Topic Segmentation. }
Long-term dialogue history comprises multiple sessions of user--assistant turns, yet neither granularity is suitable for memory construction~\cite{pan2025memory,tan2025prospect}. Individual turns often lack sufficient context. A full session may contain several interleaved topics, increasing routing ambiguity when semantically distinct spans require different memory representations. \shortname~accordingly adopts topic segments as the basic construction granularity.

Within each session, let $u_i$ and $u_{i+1}$ denote adjacent retained utterances, and let $s_i=\mathrm{sim}(u_i,u_{i+1})$ denote their cosine similarity. Topic transitions are typically accompanied by abrupt decreases in adjacent-utterance similarity. For each valid position whose left and right windows are complete, the corresponding valley depth is computed as
\begin{equation}
d_i=
\frac{1}{w}\sum_{j=1}^{w}s_{i-j}
+
\frac{1}{w}\sum_{j=1}^{w}s_{i+j}
-
2s_i,
\end{equation}
where $w$ is the number of neighboring similarity scores considered on each side. Let $D=\{d_i\}$ collect the valley depths within the current session. The adaptive boundary threshold is defined as
\begin{equation}
\tau=
\mu_D+0.05\sigma_D,
\end{equation}
where $\mu_D$ and $\sigma_D$ are the mean and standard deviation of $D$, respectively. The computed positions are treated as candidate valleys; among them, positions that are local maxima of $d_i$ and satisfy $d_i\ge\tau$ are selected as topic boundaries. The resulting segments $S$ serve as the basic decision units for subsequent write scheduling.

\noindent\textbf{$\bullet$ Write Scheduling. }
Given a topic segment $S_i$, the write scheduler selects the lowest-cost representation that preserves the evidence required for future use. Let $\mathcal{Y}=\{\texttt{ignore},\texttt{profile},\texttt{event},\texttt{record}\}$ denote the available write actions, and let $C(y,S_i)$ and $L(y,S_i)$ denote the representation cost and potential evidence loss incurred by assigning $S_i$ to action $y$, respectively. We express the scheduling design objective as
\begin{equation}
y_i=
\arg\min_{y\in\mathcal{Y}} C(y,S_i)
\quad
\text{s.t.}
\quad
L(y,S_i)\leq\epsilon,
\end{equation}
where $\epsilon$ denotes the maximum acceptable loss of information required for reliable future reasoning. Rather than estimating $C(\cdot)$ and $L(\cdot)$ numerically, \shortname~operationalizes this objective through fixed routing criteria.

Specifically, an LLM-based scheduler evaluates each segment along three discriminative dimensions: \textit{stability}, indicating whether the information remains reusable across interactions; \textit{temporal dependence}, indicating whether its meaning depends on state changes or chronological relations; and \textit{fidelity requirement}, indicating whether accurate reuse requires preserving detailed source content. Stable and readily structured information is assigned to \texttt{profile}; updates, progress, and temporally ordered states are assigned to \texttt{event}; information-dense content that cannot be safely compressed is assigned to \texttt{record}; and content without persistent evidentiary value is assigned to \texttt{ignore}. Unlike key utterance filtering, which removes structurally low-information turns before semantic processing, \texttt{ignore} applies to meaningful topic segments that do not warrant long-term storage, such as temporary requests. The routing dimensions correspond to mutually distinguishable evidence properties rather than semantic topics, making the decision less ambiguous across heterogeneous content.

The scheduler follows a fixed decision protocol encoded in the instruction prompt. It first identifies the evidence-bearing content in $S_i$, evaluates the three routing dimensions, and then applies a predefined priority order to resolve overlapping cases: temporal dependence takes precedence over stability, while high fidelity requirements override compressive representations. It returns
\begin{equation}
r_i=
\langle
y_i,\;
I_i,\;
\xi_i
\rangle,
\end{equation}
where $y_i\in\mathcal{Y}$ is the selected write action, $I_i$ contains the source indices supporting the decision, and $\xi_i$ contains the extracted attribute--value pairs when $y_i=\texttt{profile}$ and is empty otherwise. The output is restricted to a predefined schema, grounding each decision in the original segment.

In our implementation, the scheduler uses the same instruction-following LLM backbone as the corresponding memory-system configuration, including GPT-4.1-mini in the closed-source setting and Qwen3-8B in the open-source setting. Each topic segment requires one forward call, and the scheduler returns the structured result $r_i$.

\noindent\textbf{$\bullet$ Memory Materialization. }
Given $r_i$, \shortname~recovers the supporting dialogue span $X_i=\mathcal{H}[I_i]$ and materializes it according to $y_i$:
\begin{equation}
m_i=
\begin{cases}
\varnothing,
& y_i=\texttt{ignore},
\\[2pt]
\xi_i,
& y_i=\texttt{profile},
\\[2pt]
\mathrm{LLM}(X_i;e,t,z),
& y_i=\texttt{event},
\\[2pt]
\left\langle
\mathrm{LLM}(X_i;g),\;
\mathrm{NER}(X_i),\;
I_i
\right\rangle,
& y_i=\texttt{record}.
\end{cases}
\end{equation}
For profile memory, the scheduler has already extracted $\xi_i=\{(a_{ij},v_{ij})\}_{j=1}^{n_i}$, so the system stores these structured attributes directly without an additional LLM call. For event memory, $\mathrm{LLM}(X_i;e,t,z)$ denotes a schema-constrained call that returns only
\begin{equation}
\mathrm{LLM}(X_i;e,t,z)
=
\langle
e_i,\;
t_i,\;
z_i
\rangle,
\end{equation}
where $e_i$, $t_i$, and $z_i$ denote the event topic, temporal anchor, and state description, respectively. For record memory, $\mathrm{LLM}(X_i;g)$ generates only a concise retrieval gist $g_i$, while $\mathrm{NER}(X_i)$ extracts a keyword and entity set $\kappa_i$ using the lightweight GLiNER model~\cite{gliner}; the source pointer is retained as $p_i=I_i$. The resulting record is therefore $m_i^{\texttt{record}}=\langle g_i,\kappa_i,p_i\rangle$. These constrained outputs avoid free-form summarization: profile memory reuses structured extraction, event memory retains only temporal state fields, and record memory stores a lightweight retrieval index while preserving access to the original dialogue. The resulting heterogeneous memory bank is
\begin{equation}
\mathcal{M}
=
\mathcal{M}^{\texttt{profile}}
\cup
\mathcal{M}^{\texttt{event}}
\cup
\mathcal{M}^{\texttt{record}}.
\end{equation}

\subsection{Selective Memory Evolution}
Different memory representations require distinct maintenance strategies. Profile memory stores stable user attributes, while record memory preserves immutable source-grounded indices. In contrast, event memory represents ongoing processes refined through later interactions. Therefore, \shortname~restricts evolution to event memory, avoiding unnecessary consolidation of other memory types.

To reduce online maintenance cost, \shortname~decouples memory writing from evolution through deferred updates. Newly constructed event memories are accumulated in an event buffer, and consolidation is triggered only when the buffer reaches a predefined capacity. By amortizing multiple updates into one evolution step, this strategy reduces expensive LLM invocations while keeping newly written memories immediately retrievable.

During evolution, each buffered event retrieves related event memories based on their topics and states to form a localized context. The retrieved events are ordered by temporal anchors, allowing the LLM to infer the latest state and update the corresponding memory incrementally rather than reconstructing it. An incoming event may add evidence to an existing topic, revise its state, or extend its timeline. Event memory therefore evolves through localized updates while avoiding repeated global consolidation.

\subsection{Adaptive Evidence Composition}
Given an incoming query $q$, \shortname~first identifies its evidence requirements instead of retrieving a fixed number of memories. An LLM planner analyzes evidence granularity, temporal dependence, fidelity, and aggregation scope, and follows a fixed prompt template to produce retrieval plan
\begin{equation}
\pi_q=
\left\langle
\mathcal{C}_q,\;
d_q,\;
\mathcal{M}_q,\;
\mathbf{w}_q,\;
\mathbf{k}_q
\right\rangle,
\end{equation}
where $\mathcal{C}_q$ contains entity, attribute, temporal, and relational constraints; $d_q$ describes the evidence demand; $\mathcal{M}_q\subseteq\{\texttt{profile},\texttt{event},\texttt{record}\}$ denotes the required memory types; $\mathbf{w}_q=\{w_q^c\}_{c\in\mathcal{M}_q}$ represents the query's relative dependence on each selected memory type, with $\sum_{c\in\mathcal{M}_q}w_q^c=1$; and $\mathbf{k}_q=\{k_q^c\}_{c\in\mathcal{M}_q}$ specifies the retrieval depth for each type. Stable attribute lookup places greater dependence on profile memory, temporal change and state tracking emphasize event memory, and detail-sensitive verification or source traceback emphasizes record memory. Query-specific evidence demands determine the retrieval budget allocated to each selected memory type.

Guided by $\pi_q$, \shortname~retrieves the top-$k_q^c$ candidates from each selected memory type $c\in\mathcal{M}_q$. Profile retrieval matches structured attributes, event retrieval considers topics and temporal states, and record retrieval uses retrieval gists. The retrieved candidates are then reranked as
\begin{equation}
\mathrm{score}(q,m)
=
w_q^c
\left[
\mathrm{Rel}_c(q,m)
+
\mathrm{Match}(\mathcal{C}_q,m)
\right],
\end{equation}
where $\mathrm{Rel}_c(q,m)$ measures type-specific relevance, $\mathrm{Match}(\mathcal{C}_q,m)$ measures consistency with the query constraints, and $w_q^c$ reflects the query's dependence on memory type $c$. This reranking favors both relevant memories and evidence containing critical constraints or temporal anchors.

Finally, \shortname~composes the retrieved memories according to their evidence roles rather than concatenating a global top-$k$ list. Profile entries provide stable user context, event entries are chronologically ordered to expose state transitions, and record pointers are expanded to the original dialogue spans only when detailed evidence is required. The resulting evidence is assembled within the available context budget for downstream reasoning and answer generation.
\begin{table*}[t]
\centering

\begingroup
\footnotesize
\renewcommand{\arraystretch}{1.08}
\setlength{\tabcolsep}{2.8pt}
\setlength{\aboverulesep}{0.7pt}
\setlength{\belowrulesep}{0.7pt}
\setlength{\cmidrulekern}{1.5pt}

\begin{tabular*}{\textwidth}{
@{\extracolsep{\fill}}
ll|cc|ccc|cc|ccc
@{}
}
\toprule
\multirow{3}{*}{\textbf{Dataset}}
&
\multirow{3}{*}{\textbf{Method}}
&
\multicolumn{5}{c|}{\textbf{GPT-4.1-mini}}
&
\multicolumn{5}{c}{\textbf{Qwen3-8B}}
\\[-2pt]

\cmidrule(lr){3-7}
\cmidrule(l){8-12}

&
&
\multicolumn{2}{c|}{\textbf{Effectiveness $\uparrow$}}
&
\multicolumn{3}{c|}{\textbf{Efficiency $\downarrow$}}
&
\multicolumn{2}{c|}{\textbf{Effectiveness $\uparrow$}}
&
\multicolumn{3}{c}{\textbf{Efficiency $\downarrow$}}
\\[-2pt]

\cmidrule(lr){3-4}
\cmidrule(lr){5-7}
\cmidrule(lr){8-9}
\cmidrule(l){10-12}

&
&
\textbf{Recall}
&
\textbf{Acc.}
&
\textbf{Build}
&
\textbf{Infer.}
&
\textbf{Latency}
&
\textbf{Recall}
&
\textbf{Acc.}
&
\textbf{Build}
&
\textbf{Infer.}
&
\textbf{Latency}
\\
\midrule

\multirow{7}{*}{LoCoMo}
&
FullText
& 64.62 & 56.92
& -- & 16.91 & 5.78
& 56.26 & 50.83
& -- & 16.91 & 4.77
\\

&
Naive RAG
& 57.68 & 64.09
& -- & 0.60 & 1.65
& 51.61 & 51.04
& -- & 1.05 & 1.21
\\

\cline{2-12}

&
MemoryOS
& 41.62 & 65.39
& 386.40 & 5.03 & 3.62
& 51.23 & 63.04
& 429.03 & 5.73 & 4.97
\\

&
A-Mem
& 80.08 & 67.58
& 1197.28 & 22.02 & 7.30
& 78.74 & 69.80
& 1172.94 & 25.31 & 8.84
\\

&
LightMem
& \underline{82.17} & \underline{79.33}
& 106.67 & \underline{3.98} & 4.88
& \underline{79.08} & \underline{78.57}
& 129.62 & \underline{3.78} & 4.26
\\

&
SimpleMem
& 75.58 & 76.86
& \underline{94.90} & 5.54 & \underline{3.20}
& 73.40 & 71.14
& \underline{119.45} & 9.39 & \underline{4.07}
\\

&
\textbf{\shortname}
& \textbf{83.80}${^*}$ & \textbf{84.87}${^*}$
& \textbf{69.45}${^*}$ & \textbf{3.04}${^*}$ & \textbf{2.77}${^*}$
& \textbf{85.55}${^*}$ & \textbf{84.41}${^*}$
& \textbf{92.92}${^*}$ & \textbf{3.01}${^*}$ & \textbf{3.23}${^*}$
\\

\midrule

\multirow{7}{*}{LongMemEval-S}
&
FullText
& 100.00 & 77.40
& -- & 112.50 & 9.92
& -- & --
& -- & -- & --
\\

&
Naive RAG
& 81.07 & 67.60
& -- & 11.93 & 7.54
& 63.38 & 65.00
& -- & 5.33 & 6.03
\\

\cline{2-12}

&
MemoryOS
& 80.80 & 67.80
& 669.20 & 9.19 & 5.42
& 75.40 & 63.40
& 720.27 & 9.73 & 5.87
\\

&
A-Mem
& 92.83 & 71.40
& 1330.77 & 15.46 & 4.17
& 84.95 & 69.00
& 1352.68 & 16.24 & 4.53
\\

&
LightMem
& 90.18 & \underline{76.73}
& \underline{150.39} & \underline{3.78} & 3.93
& 89.77 & 69.80
& \textbf{157.60} & \underline{4.56} & 3.12
\\

&
SimpleMem
& \underline{95.15} & 74.60
& 162.38 & 9.17 & \underline{3.46}
& \textbf{95.32} & \underline{74.60}
& 174.31 & 14.60 & \underline{2.67}
\\

&
\textbf{\shortname}
& \textbf{97.67}${^*}$ & \textbf{91.80}${^*}$
& \textbf{117.61}${^*}$ & \textbf{3.62}${^*}$ & \textbf{2.16}${^*}$
& \underline{93.30} & \textbf{77.40}${^*}$
& \underline{160.35} & \textbf{3.95}${^*}$ & \textbf{2.14}${^*}$
\\

\bottomrule
\end{tabular*}
\caption{
Overall effectiveness and efficiency on LoCoMo and LongMemEval-S.
Recall and Accuracy are averaged over 5 runs; \(^{*}\) denotes a significant improvement over the strongest memory-based baseline under paired testing (\(p<0.05\)).
Build tokens are averaged per conversation, whereas inference tokens and latency are averaged per question.
Tokens are reported in thousands (K) and latency in seconds (s).
Best and second-best memory-based results are \textbf{bolded} and \underline{underlined}; FullText and Naive RAG are references only.
}
\label{tab:overall_effectiveness_efficiency}
\endgroup
\end{table*}

\section{Experiments}
\subsection{Experimental Setup}
\noindent\textbf{$\bullet$ Benchmark Datasets.}
We evaluate \shortname~on two long-term conversational question answering benchmarks: \textbf{LoCoMo}~\cite{locomo} and \textbf{LongMemEval-S}~\cite{longmemeval}. LoCoMo contains 10 conversations averaging 16K tokens and up to 32 sessions. Since Category 5 contains adversarial/unanswerable questions without gold answers, we evaluate only Categories 1--4. LongMemEval-S contains 500 interaction histories averaging 115K tokens across 40--50 sessions, each paired with one question.

\noindent\textbf{$\bullet$ Baselines and Backbone Models.}
We compare \shortname~with six baselines: FullText, Naive RAG, MemoryOS~\cite{memoryos}, A-MEM~\cite{A-mem}, LightMem~\cite{lightmem}, and SimpleMem~\cite{simplemem}; further details are provided in Appendix~B. We instantiate all methods with the proprietary GPT-4.1-mini and locally deployed Qwen3-8B.

\noindent\textbf{$\bullet$ Implementation Details.}
FullText directly uses the complete dialogue history, while Naive RAG retrieves the top-10 dialogue turns.
We use \textit{all-MiniLM-L6-v2} to encode memories for indexing and retrieval, and additional implementation details are provided in Appendix~C.

\noindent\textbf{$\bullet$ Evaluation Metrics. }
We evaluate all methods in terms of effectiveness and efficiency. We report LLM-judged Accuracy using GPT-4.1-mini. Since the original LoCoMo metrics, such as F1, cannot adequately assess semantic correctness, we follow the evaluation protocol of SimpleMem~\cite{simplemem} for all LoCoMo experiments. The full evaluation prompt is provided in Appendix~D.We also report \textit{Recall} for evidence retrieval. 
Efficiency is measured by construction tokens, inference tokens, and latency.

\begin{table}[t]
\centering
\renewcommand{\arraystretch}{1.05}
\setlength{\tabcolsep}{3.2pt}
\resizebox{\columnwidth}{!}{
\begin{tabular}{@{}l|cc|cc@{}}
\toprule
\multirow{2}{*}{Configuration}
&
\multicolumn{2}{c|}{LoCoMo}
&
\multicolumn{2}{c}{LongMemEval-S}
\\
\cmidrule(lr){2-3}
\cmidrule(l){4-5}
&
\textbf{Acc. }$\uparrow$
&
\textbf{Tok./Lat. } $\downarrow$
&
\textbf{Acc. }$\uparrow$
&
\textbf{Tok./Lat. } $\downarrow$
\\
\midrule

\shortname
& 84.87
& 72.49/2.77
& 91.80
& 121.23/2.16
\\
\midrule

w/o Utterance Filter
& 75.58
& 109.75/4.70
& 82.20
& 149.42/3.48
\\

w/o Topic Segment
& 82.79
& 92.92/3.98
& 79.60
& 184.33/3.96
\\

w/o Storage Schedule
& 72.79
& 123.56/3.36
& 71.00
& 172.73/3.60
\\

w/o Mem Evolution
& 84.42
& 89.38/3.43
& 81.20
& 125.14/3.42
\\

w/o Retrieval Plan
& 77.92
& 94.15/3.79
& 81.20
& 135.25/2.92
\\

\bottomrule
\end{tabular}
}
\caption{Ablation study on LoCoMo and LongMemEval-S. Token usage is reported in K tokens and latency in seconds.}
\label{tab:ablation}
\end{table}

\subsection{Overall Performance Analysis}
\label{sec:overall_analysis}

Table~\ref{tab:overall_effectiveness_efficiency} compares effectiveness and efficiency on LoCoMo and LongMemEval-S. Across both datasets and backbones, \shortname~achieves the best Accuracy with the lowest or near-lowest construction and inference costs. FullText and Naive RAG are included only as references. FullText cannot run on LongMemEval-S with Qwen3-8B because the average 115K-token history exceeds its context window.

\noindent\textbf{$\bullet$ Joint effectiveness--efficiency advantage. }
Compared with the strongest memory-based baselines, \shortname~improves Accuracy by 5.54/5.84 points on LoCoMo and 15.07/2.80 points on LongMemEval-S under GPT-4.1-mini/Qwen3-8B. Meanwhile, it reduces LoCoMo construction cost by 26.8\%/22.2\% and requires only 117.61K/160.35K construction tokens on LongMemEval-S. Unlike A-Mem, which consumes over 1.2M construction tokens, \shortname~achieves stronger effectiveness through heterogeneous memory construction and query-adaptive evidence composition.

\noindent\textbf{$\bullet$ Dataset-specific observations. }
LoCoMo contains 1,540 answerable questions over relatively shorter conversations and covers diverse reasoning types. It therefore emphasizes flexible access to heterogeneous evidence. With GPT-4.1-mini, \shortname~achieves 83.80 Recall and 84.87 Accuracy, improving the strongest baselines by 1.63 and 5.54 points, while reducing construction tokens, inference tokens, and latency by 26.8\%, 23.6\%, and 13.4\%, respectively. This benefits from representing stable facts, evolving states, and detail-intensive content as profile, event, and record memories, and selecting them according to each query.

LongMemEval-S instead contains much longer histories, averaging approximately 115K tokens across dozens of sessions, and places greater emphasis on long-range evidence preservation and cross-session state tracking. With GPT-4.1-mini, \shortname~reaches 97.67 Recall and 91.80 Accuracy, exceeding the strongest baselines by 2.52 and 15.07 points, while using only 117.61K construction tokens, 3.62K inference tokens, and 2.16,s latency. The larger Accuracy gain highlights the value of selectively evolving event memories and retrieving only the evidence required from raw histories.

\noindent\textbf{$\bullet$ Robustness across backbone LLMs. }
Qwen3-8B exhibits the same overall trend as GPT-4.1-mini, with \shortname~maintaining the highest Accuracy and lowest inference latency on both datasets. This consistency indicates that the gains arise primarily from the memory design rather than a particular backbone.

\begin{figure}[t]
    \centering
    \includegraphics[
        width=\linewidth
    ]{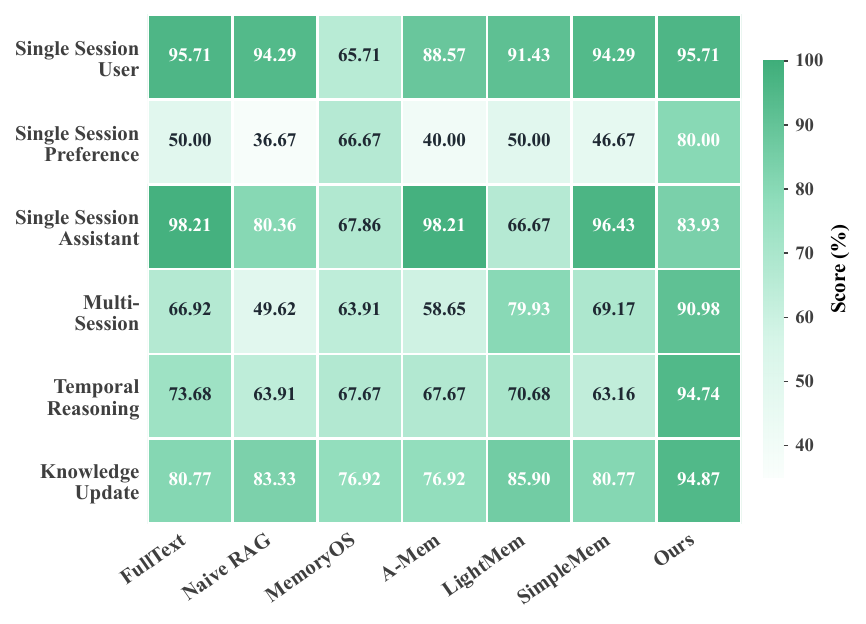}
    \caption{Accuracy of GPT-4.1-mini across different question categories on LongMemEval-S.}
    \label{fig:long}
\end{figure}

\begin{figure}[t]
    \centering
    \includegraphics[
        width=\linewidth
    ]{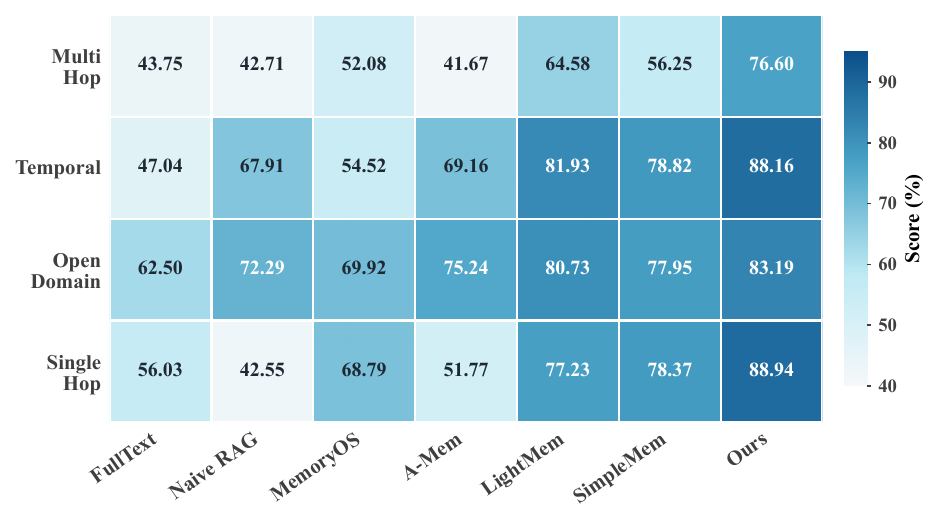}
    \caption{Accuracy of GPT-4.1-mini across different question categories on LoCoMo.}
    \label{fig:locomo}
\end{figure}

\subsection{Performance across Question Types}
Fig.~\ref{fig:long} and~\ref{fig:locomo} report the Accuracy of different methods across question types using GPT-4.1-mini.

\noindent\textbf{$\bullet$ LoCoMo. }
\shortname~achieves the highest Accuracy across all four answerable question types, demonstrating consistent benefits under different evidence demands. Compared with the strongest baseline for each category, it improves Multi-Hop, Temporal, Open-Domain, and Single-Hop Accuracy by 12.02, 6.23, 2.46, and 10.57 points, respectively. The large gains on Single-Hop and Open-Domain questions indicate that profile and source-grounded record memories preserve both reusable facts and fine-grained details. Meanwhile, the improvements on Temporal and Multi-Hop questions show that event memory and Adaptive Evidence Composition can organize evolving states and combine complementary evidence across memory types.

\noindent\textbf{$\bullet$ LongMemEval-S.}
The gains on LongMemEval-S are concentrated in question types that require long-range state tracking and cross-session evidence integration. \shortname~improves significantly over the strongest memory-based baseline on Single Session Preference, Multi-Session, Temporal Reasoning and Knowledge Update. These results directly support Selective Memory Evolution, which incrementally maintains temporally changing information, and Adaptive Evidence Composition, which retrieves and orders relevant evidence from long histories.

\subsection{Ablation Study}
We assess the contribution of each key component in \shortname~ using the following variants:
(1) \textbf{w/o Key Utterance Filtering}: performs topic segmentation over all dialogue turns rather than filtering for topic-advancing user utterances;
(2) \textbf{w/o Topic Segmentation}: treats each session as a single dialogue unit for memory construction;
(3) \textbf{w/o Memory Storage Scheduling}: replaces scheduling and heterogeneous memory representations with a unified summary-based strategy;
(4) \textbf{w/o Event Memory Evolution}: stores newly constructed event memories without topic-level consolidation or incremental evolution; and
(5) \textbf{w/o Retrieval Planning}: replaces demand-aware evidence composition with standard top-$k$ embedding retrieval.

As shown in Table~\ref{tab:ablation}, all components contribute to the effectiveness or efficiency of \shortname.
\textbf{1) Key utterance filtering.}
Removing key utterance filtering introduces non-informative turns, leading to lower accuracy and higher token usage and latency.
\textbf{2) Topic segmentation.}
Without topic segmentation, session-level inputs conflate distinct discussions, resulting in coarser construction units, higher costs, and severe degradation on LongMemEval-S.
\textbf{3) Memory storage scheduling.}
Replacing heterogeneous memory representations with unified summaries causes the largest accuracy drop on both datasets and increases token usage, confirming that minimal sufficient representations are central to evidence preservation and efficiency.
\textbf{4) Event memory evolution.}
Removing event memory evolution substantially degrades LongMemEval-S performance. Without consolidation, related event states remain fragmented, requiring more evidence during inference and offsetting the saved evolution cost.
\textbf{5) Retrieval planning.}
Replacing retrieval planning with fixed top-$k$ retrieval reduces accuracy and increases inference cost, highlighting the importance of query-specific memory selection and retrieval budgets.

\subsection{Case Study}

\begin{figure}[!t]
    \centering
    \includegraphics[
        width=\columnwidth,
        height=0.82\textheight,
        keepaspectratio
    ]{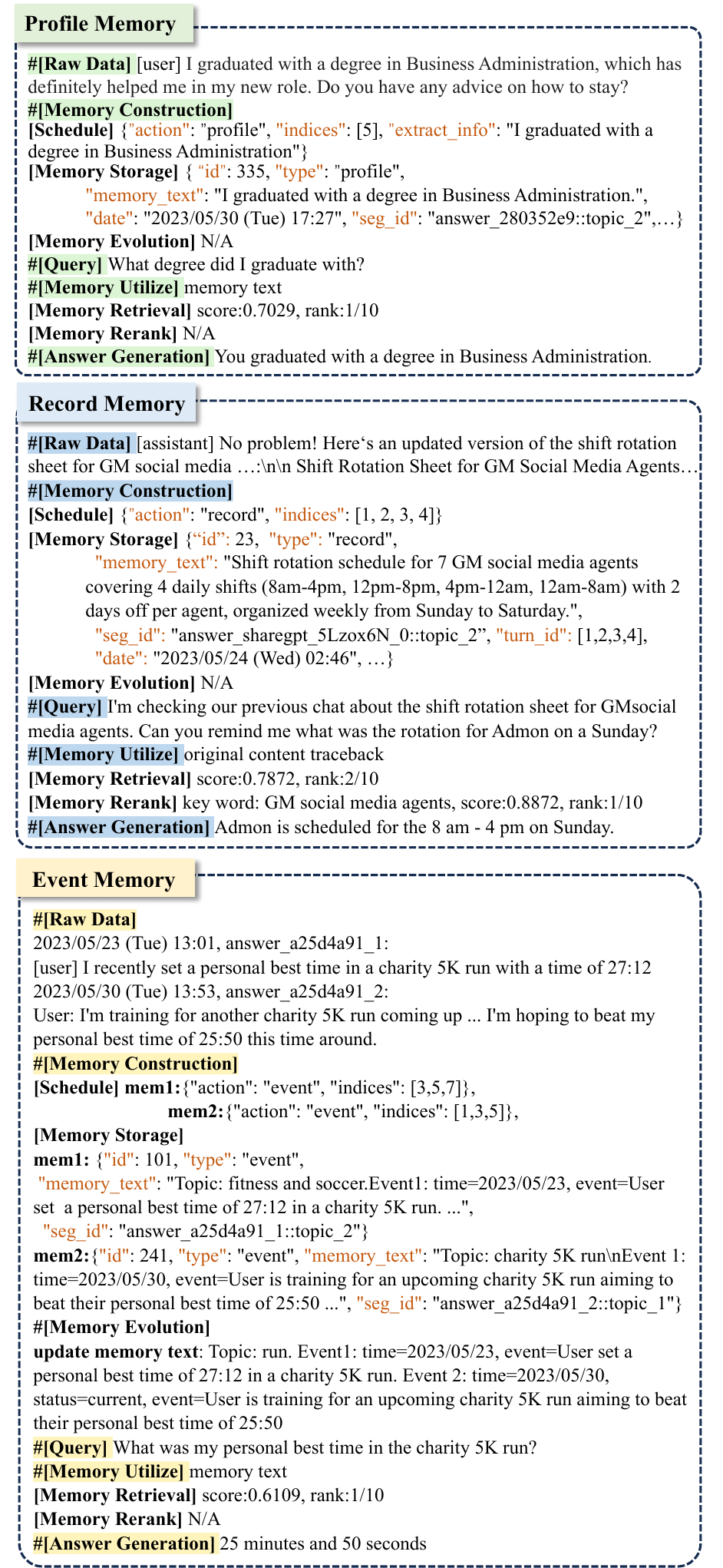}
    \caption{Case study for the construction and utilization process of profile memory, record memory and event memory.}
    \label{fig:case123}
    \vspace{-5pt}
\end{figure}

Fig.~\ref{fig:case123} illustrates how \shortname~identifies different types of conversational information, stores them as heterogeneous memory representations, and selectively evolves dynamic event states. These memories provide targeted evidence for future queries.

\noindent\textbf{$\bullet$ Profile Memory.}
The top panel of Fig.~\ref{fig:case123} shows how stable and reusable information, such as user profiles and preferences, is identified during memory scheduling and stored as compact profile memory in a minimally sufficient form without subsequent evolution. For future queries about user facts, profile memory provides precise evidence without introducing irrelevant conversational context.

\noindent\textbf{$\bullet$ Record Memory.}
The middle panel shows how information-rich conversational content is stored as record memory. \shortname~uses a lightweight local model to generate keywords while preserving pointers to the original dialogue, avoiding information loss caused by excessive abstraction. During retrieval, the reranking mechanism uses these keywords to better align retrieved records with the query, enabling more accurate evidence selection.

\noindent\textbf{$\bullet$ Event Memory.}
The bottom panel shows how evolving event states are consolidated into coherent temporal evidence. \shortname~summarizes each event into a compact topic representation with temporal anchors, and selectively evolves related memories by organizing their state changes into a chronological timeline. For future queries about event updates, these evolved memories provide coherent temporal evidence for accurate question answering.

\section{Conclusion}
In this paper, we study how long-term memory systems should represent heterogeneous conversational information. We argue that uniform compression is mismatched with dialogue history: stable attributes can be compactly structured, evolving information requires temporal state tracking, and detail-intensive content should preserve access to its source. Based on this insight, we propose \shortname, which writes information into profile, event, and source-grounded record memories, selectively evolves changing events, and composes evidence according to query-specific demands. This design avoids redundant processing of easily compressible information and irreversible loss of fidelity-sensitive evidence. Experiments on LoCoMo and LongMemEval-S demonstrate consistently stronger answer quality with lower construction and inference costs.


\bibliography{aaai2027}


\end{document}